\pdfoutput=1
\documentclass[11pt]{article}

\usepackage{acl2023}
\usepackage{xurl}
\usepackage{times}
\usepackage{latexsym}

\usepackage[T1]{fontenc}
\usepackage[utf8]{inputenc}

\usepackage{microtype}
\usepackage{graphicx}
\title{Ethical Considerations for Machine Translation of Indigenous Languages: Giving a Voice to the Speakers}

\author{ \textbf{Manuel Mager${ }^{\heartsuit}$\thanks{~~Work done while at  the University of Stuttgart.}\quad
Elisabeth Mager${ }^{\sharp}$ }\\
\textbf{Katharina Kann${ }^{\spadesuit}$
Ngoc Thang Vu${ }^{\diamondsuit}$  } \\
${ }^{\heartsuit}$AWS AI Labs \quad  \quad
${ }^{\sharp}$Universidad Nacional Autónoma de México \\
${ }^{\spadesuit}$University of Colorado Boulder \quad ${ }^{\diamondsuit}$University of Stuttgart}

\date{}
\begin{document}
\maketitle
\begin{abstract}
In recent years machine translation has become very successful for high-resource language pairs. This has also sparked new interest in research on the automatic translation of low-resource languages, including Indigenous languages. 
However, the latter are deeply related to the ethnic and cultural groups that speak (or used to speak) them. The data collection, 
modeling and deploying machine translation systems thus result in new ethical questions that must be addressed. 
Motivated by this, we first survey the existing literature on ethical considerations for the documentation, translation, and general natural language processing for Indigenous languages. Afterward, we conduct and analyze an interview study to shed light on the positions of community leaders, teachers, and language activists regarding ethical concerns for the automatic translation of their languages. Our results show that the inclusion, at different degrees, of native speakers and community members is vital to performing better and more ethical research on Indigenous languages.
\end{abstract}

\section{Introduction}
\label{sec:introduction}

With the advancement of data-driven machine translation (MT) systems, it has become possible to, with varying degrees of quality, to translate between any pair of languages. The only precondition is the availability of enough  monolingual \cite{lample2018unsupervised,artetxe2018unsupervised} or parallel data \cite{vaswani2017attention,bahdanau2015neural}. There are many advantages to having high-performing MT systems. For example, they increase access to information for speakers of indigenous languages \cite{mager-etal-2018-challenges} and can assist revitalization efforts for these languages \cite{zhang2022can}. 

Research on machine translation as well as natural language processing (NLP) more generally is moving towards low-resourced setups and multilingual models. Thus, the NLP community needs to open the discussion of repercussions and best practices for research on indigenous languages (that in most cases are also low-resourced) since non-artificial languages cannot exist without a community of people that use (or have traditionally used) them to communicate.

Indigenous languages further differ from more widely used ones in a crucial way: they are commonly spoken by small communities, and many communities use their language (besides other features) as a delimiter to define their own identity \cite{palacios2008lengua,enriquez2019rol}, and have in many cases also a certain degree of endangerment. Furthermore, in some cases, highly sensitive information -- such as secret aspects of their religion -- has been encoded with the help of their language \cite{carlos2016richard}. This is why, in recent years, discussions around ethical approaches to studying endangered languages have been started \cite{smith2021decolonizing,liu2022not}. When we consider the past (and present) of some of the communities that speak these languages, we will find a colonial history, where research is not the exception \cite{bird-2020-decolonising}.
Therefore,  it is possible to trespass on ethical limits when using typical NLP and data collection methodologies \cite{dwyer2006ethics}. 

In this work, we explore the basic concepts of ethics related to MT of endangered languages with a special focus on Indigenous communities, surveying previous work on the topic. To better understand the expectations and concerns related to the development of MT systems for Indigenous communities, we then conducted an interview study with 22 language activists, language teachers, and community leaders who are members of Indigenous communities from the Americas. Additionally, we also performed 1:1 dialogues with two study participants to deepen our understanding of the matter. The goal is to answer the following research questions: \textit{How do community members want to be involved in the MT process, and why?} \textit{Are there sensible topics that are not ethical to translate, model, or collect data without the community's explicit permission?} \textit{How can we collect data in an ethical way?}

Surprisingly, most survey participants positively view MT for their languages. However, they believe research on their languages should be done in close collaboration with community members. Open access to research discoveries and resources is also valued highly, as well as the high quality of the resulting translations. The personal interviews also confirmed this.
Thus, our most important finding is that it is crucial to work closely with the communities to understand delicate ethical topics when developing MT systems for endangered languages.

A Spanish translation of this paper is included in Appendix \ref{sec:spanish}. This translation aims to share our findings with all study participants and their communities and facilitate access to a broader audience in the Americas.

\section{Defining ``Endangered Language''}
\label{sec:endangered}

Terms frequently used in NLP are \textit{low-resource language}, \textit{resource-poor language}, and \textit{low-resource setting}. Those terms are not highlighting the fact that many low-resource languages are also endangered \cite{liu2022not}. Instead, they emphasize the critical machine learning problem of getting a data-driven approach to perform well with a smaller-than-ideal amount of available data (or just fewer data than what has been used for other languages). In this case, algorithmic or technological innovations are needed to close the performance gap between high-resource and resource-poor languages. This further implies that being low-resourced is not a property of a language but a term that only makes sense in the context of a particular task or tasks.

In contrast, the term \textit{endangered language} refers to a language with a certain degree of danger for its existence.\footnote{In this paper, we will discuss only non-artificially created languages.} 
 Endangered languages are relevant for our study, as most Indigenous languages are also endangered \cite{hale1992endangered}. 
According to the UNESCO classification, \cite{moseley2010atlas} languages can be sorted into the following different categories:

\begin{itemize}
    \item \textit{safe}: spoken by all generations;
    \item \textit{vulnerable}: restricted just to a certain domain (e.g., inside the family);
    \item  \textit{definitely endangered}: it has no kids that  speak the language;
    \item \textit{severely endangered}: only elder people speak it;
    \item \textit{critical endangered}: there are only speakers left with partial knowledge, and they use it infrequently;
    \item \textit{extinct}, when there are no persons able to speak the language anymore. 
\end{itemize}  

Languages can become endangered due to social, cultural, and political reasons; most commonly conquests and wars, economic pressures, language policies from political powers, assimilation of the dominant culture, discrimination, and language standardization \cite{austin2013endangered}. As we can see, the problem of how a language gets endangered involves factors that must be addressed in the ethical approach of any study. 
On the machine learning side, an additional challenge arises: data for endangered languages is not easily available (or, in fact, available at all), as these languages have limited media production \citep[TV shows, literature, internet blogs; ][]{hamalainen2021endangered}. One possible source of data for these languages is already existing documents in form of books, records, and archives \cite{bustamante-etal-2020-data}.

\section{Ethics and MT}
\label{sec:theory}
\subsection{Ethics and Data}
\label{subsec:ethics_documentation}

The study of endangered languages in indigenous communities has a long history, with the most prominent questions being focused mainly on the challenge of data collection \cite{smith2021decolonizing}. 

One of the common forms of this is to use normative ethics (deontology). Examples of relevant guidelines include those from The Australian Institute of Aboriginal and Torres Strait Islander Studies;\footnote{\url{https://www.jstor.org/stable/pdf/26479543.pdf}} the Ethical statement of the Linguistic Society of America;\footnote{\url{https://www.linguisticsociety.org/content/lsa-revised-ethics-statement} \url{-approved-july-2019}} and the DOBES code of conduct.\footnote{\url{https://dobes.mpi.nl/ethical_legal_aspects/DOBES-coc-v2.pdf}} These lists are the results of broad discussions which have taken place over decades. In this debate also, indigenous voices were inside academia raised \cite{smith2021decolonizing}. 

But why do we have so many attempts to set up an ethical code for linguistic fieldwork? When it comes to working with human societies, there are no easy solutions for the ethical dilemmas that arise \cite{dwyer2006ethics}. Every situation requires a unique treatment and compromise. This is why, in addition to the creation of a framework which is as general as possible, the concrete application of such principles involves continued discussion.
\newcite{dwyer2006ethics} suggests documenting the ethical issues and concerns which arise during the course of a research project and the way these issues are addressed, such that other researchers can learn from the experience. While a code of conduct or principles is good, it runs the risk of introducing either overly complicated -- or even inadequate -- regulations, relegating this needed discussion. 

Overall, we can summarize those principles that appear in all suggested lists under three big themes:
\begin{itemize}
    \item \textit{Consultation, Negotiation and Mutual Understanding}. The right to consultation of Indigenous people is stipulated in convention 167 of the International Labor Organization \cite{ilo1989c169} and states that they ``have the right to preserve and develop their own institutions, languages, and cultures''. Therefore, informing the community about the planned research, negotiating a possible outcome, and reaching a mutual agreement on the directions and details of the project should happen in all cases.
    \item \textit{Respect of the local culture and involvement}. As each community has its own culture and view of the world, researchers -- as well as any governing organizations interested in the project -- should be familiar with the history and traditions of the community. Also, it should be recommended that local researchers, speakers, or internal governments should be involved in the project.
    \item \textit{Sharing and distribution of data and research}. The product of the research should be available for use by the community, so they can take advantage of the generated materials, like papers, books, or data. 
    
\end{itemize}
Some of these commonly agreed-on principles %open to an agreement 
need to be adapted to concrete situations, which might not be easy to do via a general approach. For instance, the documentation process will create data, and the ownership of this data is a major source of discussion (cf. Sections \ref{sec:study}, \ref{sec:discussion}). Here, the traditional views of the communities might contradict the juridical system of a country \cite{daes1993discrimination}. This problem does not have a simple solution and needs to be carefully considered when collecting data. 

An additional call from these sources is to decolonize research and to stop viewing Indigenous communities as sources of data, but rather as people with their own history 
\cite{smith2021decolonizing}. The current divorce between researchers and the cultural units of the communities can lead to reinforcing colonial legacy \cite{leonard2020producing}.

As a final remark, we want to discuss the common assumption that any Ethical discussion must end with a normative setup for a field.   It reduces indigenous institutions' collective to norms that allow an individual approach to the matter \cite{meza2017etica}. This would also not allow understanding the ethical questions with their own Indigenous communal cosmovision \cite{salcedo2016vivir}. Therefore, in this text, we aim to open the MT ethical debate to the NLP researchers and the Indigenous communities based on inclusion and dialog.

\subsection{Ethics and \textit{Human} Translation}
\label{subsec:ethics_translation}

For a successful translation, the inclusion of all participants is important, requiring their equal, informal, and understanding-oriented participation \cite{nissing2009grundpositionen}. 
For \newcite{rachels1986elements}, the minimum conception of morality is that when we give ``equal weight to the interests of each individual affected by one’s decision.'' The question is how authors’ intentions relate to the source culture's otherness, with their culturally-specific values \cite{chesterman2001proposal}. 
According to \newcite{doherty2016translations}, ``the translation process studies emerged to focus on the translator and the process of translation rather than on the end product,'' incorporating mixed-method designs to get objective observations. 
A well-documented example of the non-ethical misuse of translation is the application of translation as an instrument for colonial domination. The main aim of this colonialist vision was to ``civilize the savages'' \cite{ludescher2001instituciones}. For example, the summer institute of linguistics (SIL International)\footnote{SIL International describes itself as ``.. a global, faith-based nonprofit that works with local communities around the world to develop language solutions that expand possibilities for a better life. SIL's core contribution areas are Bible translation, literacy, education, development, linguistic research, and language tools.''. \url{https://www.sil.org/}} was used for this goal during the 20th century in countries with Indigenous cultures, translating the Bible and trying to provoke a cultural change\footnote{The role of SIL is controversial, and can not be summarized with one single statement. In our approach, we only refer to the role played related to cultural change. In many cases, the communities that got religious texts translated were already Christians, given previous colonization actions. However, there are also cases, where non-christian communities had Bibles and other religious texts translated into their language, with missionary aims. This triggered community divisions. For example, the translation of the religious texts to Wixarika \cite{fernandez2022libertad}. This also happened in the Community of Zoquipan (in the Mexican state of Nayarit), where Christians, using the SIL-translated Bible, triggered an internal conflict in the community (the first author is part of this community). For the interested reader, we also recommend \newcite{dobrin2009sil} introductory article.} in these communities \cite{delvalls1978instituto,errington2001colonial,carey2010lancelot}. Of course, these practices are not new and can be found throughout history \cite{gilmour2007missionaries}. It is essential to notice that non-ethical research can still deliver useful material and knowledge, e.g., for language revitalization \cite{premsrirat2003language}, but might inflict harm on the targeted community.

\subsection{Ethics and \textit{Machine} Translation}
\label{subsec:ethics_MT}

In the context of NLP research, the speakers are not directly involved when a model is trained \citep{pavlick2014language}. In contrast, the data collection processes \cite{fort2011crowdsourcing} and human evaluation \cite{couillault-etal-2014-evaluating} directly interact with the speakers and, therefore, have central importance regarding ethics. This is also true for the final translation service, which will interact with the broad public. 

Data collection is the first and most evident issue when it comes to translation. Modern neural MT systems require a large amount of parallel data to be trained optimally \cite{junczys2019microsoft}. One way to obtain data is from crowd-sourcing \cite{fort2011crowdsourcing}. However, this kind of job can be ill-paid and might constitute a problem for the living conditions of the workers \cite{schmidt2013good}.  
Also, data privacy is not trivial to handle. Systems must be able to filter sensitive information. 

The problem of encoding biases\footnote{It is also important to note the typological features that might make this challenging. One example are polysynthetic languages and languages without gender coding \cite{klavans-2018-computational}.}, like gender bias \cite{stanovsky2019evaluating}, is also an ethical concern for MT.
It is also necessary to disclose the limitations and issues with certain systems \cite{leidner-plachouras-2017-ethical}. 

NLP research can also be used as a political instrument of power, where we can observe mutual relationships between language, society, and the individual that ``are also the source for the societal impact factors of NLP'' \cite{horvath2017language}. In this way, NLP translation can be applied as an instrument to changing the culture of minorities as in traditional translation (cf. Section \ref{subsec:ethics_translation}). So colonizers used translation as means of imperial control and expropriation \cite{cheyfitz1997poetics,niranjana10siting}. The asymmetry of power is the cause of domination, where subaltern cultures being flooded with ``foreign materials and foreign language impositions'' is a real danger for minority cultures \cite{tymoczko2006translation}.
\newcite{schwartz2022primum} discuss the need to decolonize the scientific approach of the NLP community as a whole, expressing the need for researchers to be cognizant of the history and the cultural aspects of the communities which use the languages they are working with. Additionally, he proposes that our research should have an obligation to provide some benefit from our studies to the communities, an obligation of accountability (and therefore be in direct contact with their governing organizations), and an obligation of non-maleficence. The fact that many translation systems nowadays are multilingual\footnote{Multilingual systems refer in NLP to systems capable of translating a set of languages from and to English. In some cases, they are also able to translate between languages where English is not involved.} also result in more multi-cultural challenges \cite{hershcovich2022challenges}. 

Finally, we also want to highlight the importance of discussing MT systems in a text-to-text setup. The usage of text is constrained to certain topics and varies from community to community. For instance, Wixarika and Quechua, languages that are spoken across all generations, are used in a written fashion mostly in private messaging apps (like WhatsApp) but also have a  prolific Meme and Facebook publication generation\footnote{For example, Wixarika memes: \url{https://www.facebook.com/memeswixarika2019}, Quechua speaking group: \url{https://www.facebook.com/groups/711230846397383/}}. Even if a certain community does not widely adopt the written tradition, there are, at minimum legal obligations of the States towards indigenous languages. For example,  some constitutions recognize indigenous languages as national languages (e.g., Mexico and Bolivia), binding the state to the responsibility to translate all official pages, documents, laws, etc.,  to indigenous languages. This has not been implemented, and this case is a highly valuable application case for machine translation to assist human translation. However, our findings also apply to speech-to-text translation and speech-to-speech tasks that would cover all languages, even with no written tradition.

\begin{figure*}[h!]
    \centering
    \includegraphics[width=1\textwidth]{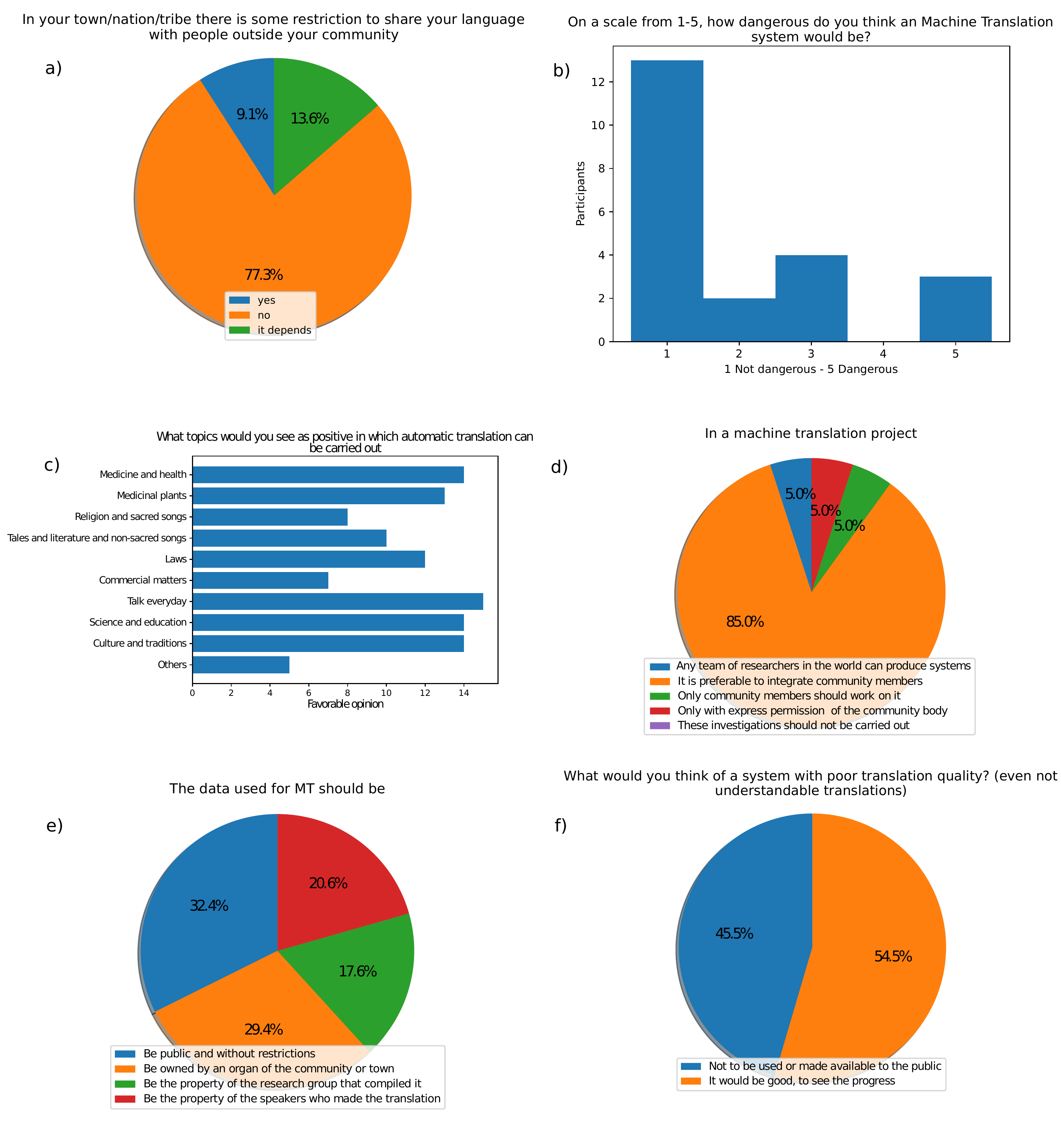}
    \caption{Study performed on 22 participants that are members of Indigenous communities from the Americas. }
    \label{fig:results}
\end{figure*}

\section{The Speakers' Opinions}
\label{sec:study}

It is important to include the opinion and vision of speakers of endangered languages in NLP research, especially for topics such as MT. Therefore, we conduct a survey study with 22 language activists, teachers, and community leaders from the Americas. Importantly, our primary goal is not only to gather quantitative input on the ethical questions regarding MT for their languages but also to collect qualitative input by asking them to expand on their answers. Additionally, we also perform an interview with a subset of two participants of the initial interview study.

\subsection{Study Design}

We focus our study on the Americas,\footnote{Different parts of the world have very different levels of wariness, not just from colonial history but precisely due to interactions with field workers.} selecting the following communities: Aymara, Chatino, Maya, Mazatec, Mixe, Nahua, Otomí, Quechua,  Tenek, Tepehuano, Kichwa of Otavalo, and Zapotec. We want to note that our study does not aim to represent a general opinion of all Indigenous tribes, nor is it a final general statement on the issue. It is a case study that surfaces the opinions of specific groups of speakers of Indigenous languages. Furthermore, the views of the interviewed individuals are their own and 
do not necessarily represent the view of their tribes, nations, or communities.

\paragraph{Quantitative and Qualitative aspects} 

For the quantitative study, we used a survey. Surveys are a well-established technique to be used with Indigenous communities with an extensive history and are used and documented by classics like Edward Tylor, Anthony Wallace, Lewis Henry Morgan. This is also true for well-recognized Mexican (Indigenous engaged) social anthropologists \cite{jimenez1985teoria,alfredo1978teoria}. 

For the qualitative part, we revisit existing positional papers and articles of Indigenous researchers and activists. Additionally, we use open questions in the survey, allowing extending the pure quantitative view to a qualitative one. Finally, we performed two 1-to-1 interviews with an activist (Mixe) and a linguist (Chatino). 

\paragraph{Participant Recruitment} We contact 
potential participants online in three ways. Our first approach is to establish communication through the potential participants' official project websites or public online accounts. This includes e-mail, Twitter, Facebook, and Instagram pages. Our second approach is to directly contact people in our target group with whom at least one of the co-authors has already established a relationship. Finally, we also published a call for participation on social media 
and check if the volunteers belong to our target group. The goals of our research, as well as the reach and data handling, are explained directly to each participant and are also included in the final form. We do not gather any personal information about the participants, like name, gender, age, etc. 
All study participants are volunteers.

\paragraph{Questionnaire} Our study consists of 12 questions. The first three questions are rather general: they ask for the tribe, nation, or Indigenous people the participant belongs to if they self-identify as an activist, community leader, or teacher, and for their fluency in their language. The remaining questions target data policies, inclusion policies, benefits and dangers of MT systems, and best research practices. 
The full questionnaire is available in the appendix. The questions are available in English and Spanish, but only one form has been filled in English, while the rest has been completed in Spanish. Therefore, the authors have automatically translated all comments which are shown in this paper. 

\subsection{Results}

The results of the study can be seen in Figure \ref{fig:results}. Additionally, we also discuss the open answers to each question to provide more insight.

\paragraph{Inclusion of Native Speakers and Permissions to Study the Language} Figure \ref{fig:results}(a) shows that 77.3\% of the participants report that their community has no restrictions regarding the sharing of their language with outside people. 
The comments for this question show that many participants are 
proud of their language and heritage: ``We are supportive and share our roots. Proud of who visits us'' We even find stronger statements against the prohibition to share: ``No one has the right to restrict the spread of the language''. 
However, there also do exist communities with restrictions. Thus, we conclude that researchers cannot assume by default that all Indigenous groups would agree to share information about their language or would be happy about research on it.

\paragraph{Benefits and Dangers of MT Systems} 
Figure \ref{fig:results}(b) shows that a strong majority of our participants think that an MT system for their language would be beneficial. However, there is also an important number of people who see at least some degree of danger. In this case, we need to look at the participants' comments to understand their worries. First, we find that a main concern for the participants is the translation quality.
The fear of inadequate translations of cultural terms is also important. In Table \ref{tab:dangers}, we can see a set of comments that illustrate these fears. One interesting comment refers to the fear of standardization of the participant's language, which could lead to a loss of diversity. In the same table, we can also see the benefits the participants expect, mostly in education and in elevating the status and usefulness of their languages. 

Table \ref{tab:topics} shows some answers to the open question on possible topics that might cause damage to the community. Most answers could not identify any possible topic that could be dangerous. However, the second most frequent answer was related to religion. Some answers are worried that ancient ceremonial secrets could be revealed. Others also show worries about the influence of Western religions. This brings us to the question if the Bible \cite{christodouloupoulos2015massively,mccarthy2020johns,agic2019jw300} is suited to use as our default corpora for MT, when an indigenous language is involved. Finally, also few answers expressed that the usage of indigenous languages in the internal organization of the community could be in danger with MT systems. In contrast, figure \ref{fig:results}(c) shows the topics that that most positive evaluation registered: everyday talks (15), science and education (14), culture and traditions (14), and medicine and health (14).

\begin{table}[h]
    \centering
    \small
    \begin{tabular}{p{0.9\linewidth}}
\bf What would you see as damaging topics that should not be machine translated? \\\hline
Anything ceremonial \\
Laws, medicine and health, science, mercantile matters, religion and sacred songs.\\
Issues that threaten organic life.\\
Western religion\\
Political situations and religions unless it is in the interest of the person.\\
Sacred songs, like those of a healer.\\

    \end{tabular}
    \caption{Some answers to the open question on possible dangers of MT for indigenous languages.}
    \label{tab:topics}
\end{table}

\begin{table}[h]
    \centering
    \small
    \begin{tabular}{p{0.9\linewidth}}
         \bf Can you think of any dangers to the language and culture, if so, which? \\\hline    
          There are cultural linguistic concepts that are only understood in our native language. \\
          The existence of so many variants would make the project little or not profitable and would lead the "experts" to an attempt to standardize language, which would be a tremendous mistake. \\
          There are cultural elements that must be taken into account. \\
          They could undoubtedly distort the proper use of the language. \\\hline
          \bf What advantages would you see with an automatic translation system? \\\hline
          The use of automatic translators in spaces such as hospitals, government offices, etc.\\
         Perhaps a contribution of modernity to the community, preservation of the native language.\\
         It would contribute to the status of indigenous languages\\
         It would contribute to the social use of our language\\
         It would facilitate teaching because you would have many support tools.
    \end{tabular}
    \caption{Open answers of speakers to questions on dangers and benefits of MT systems for their communities.}
    \label{tab:dangers}
\end{table}

\paragraph{Participation of Members of Indigenous Communities in Research } Figure \ref{fig:results}(d) shows that our study participants think it is important to include people from the targeted communities in research projects.  
This confirms the experience in linguistics, where they found a similar pattern \cite{smith2021decolonizing} (see \S\ref{subsec:ethics_documentation}). 
It is important to note that only one answer was stated that official permission is needed to perform the study.
In the comments, the right of consulting was mentioned, together with the advantages of involving community members in research: ``It is preferable [to integrate people from the community] to obtain a good system, and not just to have approximations, because only the members of the culture know how the language is used."; ``So that the vocabulary is enriched and some words that do not exist are not imposed.''; ``Carry out activities where the community can be involved, win-win.''.

\paragraph{Data Usage and Translation Quality} Regarding data ownership and accessibility, we find diverse sets of responses. First, Figure \ref{fig:results}(e) shows many different opinions. %Comparing the percentages of the answers with the comments, 
Overall, we can say that a strong feeling exists that data should be publicly available.
However, when it comes to the property of the data, opinions are more diverse. Surprisingly, an important number of participants ($17\%$) think that the external research group should own the data. Nevertheless, a higher number of participants think that the data should be owned by the community ($29.4\%$), and 20.6\% thinks it should be owned by the speakers who participate in the research. This is a difficult topic, as traditional norms and modern law systems interact (cf. Section \ref{subsec:ethics_documentation}). In the comments, we find sad examples of mistrust in academic institutions. For example, one comment talks about previous problems of their tribe, as recordings and other material taken by linguists is not accessible to them: ``Wary of academic institutions since we currently have issues accessing recordings that belong to academics and libraries and are not publicly accessible.'' However, in general, we see a wide range of opinions: ``The work of the few who take linguistic identity seriously should be valued'', 
``It could be public but always with the endorsement and consent of the community.'' This diversity demonstrates that there is a need for researchers to have a close relationship with the communities to understand the background and the aims of each particular case.

As discussed above, the quality of the final system is an important concern for many participants. In Figure \ref{fig:results}(f) we can see that publishing an experimental MT system is also controversial. The possibility of using an experimental system is liked by $54.8\%$ of our participants, which is slightly higher than the number of participants who are against this ($45.5\%$). Some opinions against it are in line with earlier worries about incorrect translations of cultural content: ``Something that is devoid of structure and cultural objectivity %is not 
cannot be made available to the public'' and ``...damage would be caused to the language and its representatives since the learners would learn in the wrong way.'' Most people with a positive opinion agree that an initially poor system could be improved over time: ``If it could be improved and corrected, that would be excellent.''

\section{Discussion}
\label{sec:discussion}

In Section \ref{sec:theory} we survey the ongoing debate on ethics in documentation, translation, and MT, before presenting an interview study in Section \ref{sec:study}. Now we discuss some the most important issues we have identified in the last section in more depth.

\paragraph{Need for Consultations with Communities } Previous experiences \cite{bird-2020-decolonising,liu2022not} as well our study highlight the need for consultation with Indigenous communities when performing research involving their languages\footnote{An example of a community engaged fieldwork is \newcite{czaykowska2009research}}. In some cases, the minimal expressed requirement is to inform speakers about new technological advances. Feedback and quality checks are also crucial for MT systems and important to the members of the communities. This consultation should include intercultural dialog as it has been a central instrument in the decision-making of indigenous communities \cite{beauclair2010eticas}. We recommend doing this by integrating community members into the loop while, of course, giving them the credit they deserve. 

\paragraph{Legal systems vs. Traditional Views of Communal Knowledge Ownership} Legal systems and, with that, copyright laws vary by country. However, legal rules are sometimes in conflict with the traditional views of Indigenous people \cite{dwyer2006ethics}. Thus, when working with Indigenous communities, we recommend discussing and agreeing upon ownership rights with annotators or study participants prior to starting the work to find an arrangement everyone is happy with. We would also like to point out that, according to our case study, a general feeling is that data and research results need to be accessible to the community speaking the language. This contradicts the practice of some documentation efforts that close the collected data to the public and even to the speakers of the community \cite{Heriberto2021Nuevas}. Some participants in our study even suggest the usage of Creative Commons (CC)\footnote{\url{https://creativecommons.org/licenses/}} for data.  However, the use of CC might not be the best licensing option, as it not designed specifically for the needs of Indigenous. 
Finally, whenever collected data are used for commercial usage, special agreements involving financial aspects are crucial. 
\paragraph{Permissions } Some communities require that a permit from their governing entity be obtained when someone, not a member, wants to study their language.  This might be difficult as sometimes there is no central authority. Figuring out from whom to get permission can be challenging in such scenarios.  However, as we see in this study, many communities do not require this permission.  A promising project that aims to simplify this topic is the KP labels\footnote{
\url{https://localcontexts.org/labels/traditional-knowledge-labels/}}. It is a set of labels that communities can use to express their permissions and willingness to cooperate with researchers and external projects.

\paragraph{Personal Data} From the free-text answers, we further learn that, for many speakers, using their own language in their daily environment helps them protect their privacy: 
Their conversations can only be understood by their family or close environment. This concern of data handling is, however, also valid for other languages.

\paragraph{Concerns about  
Private Information of the Community} The previous point can further be extended to assemblies and other organizational meetings, where the language barrier is used to keep their decisions or strategies private. This is one worry that the communities have with MT and the possible topics that might be harmful for them. Some communities also have general concerns about sharing their language with people that do not belong to them (e.g., the Hopi Dictionary controversy \cite{hill2002publishing}). For this case, it is important not to approach this issue from a Western legal point of view and go towards traditional internal governance practices and norms and consultation with the communities. 

\paragraph{Religion and the Bible } Regarding problematic domains for MT, multiple survey participants mentioned religion. This is quite relevant for the NLP community, as the broadest resource currently available for minority languages is the Bible. As seen in Section \ref{subsec:ethics_translation}, the colonial usage of translation of religious texts \cite{niranjana1990translation} is precisely the origin of these detests. Thus, we recommend that NLP and MT researchers use the Bible carefully, through a consultation process, and consider its impacts. Nevertheless, without a close relationship with each community (e.g., in a massive multilingual MT experiment), the recommendation is to void using the Bible.

\paragraph{Technology and data Sovereignty} Having technology for their own languages is well seen by most study participants. However, we also find a strong wish to participate directly in the development of MT systems. This requires more inclusion of Indigenous researchers in NLP. Therefore, training Indigenous researchers and engineers is an important task that we recommend should be valued more highly by the NLP and MT communities. We are aware that existing inequalities cannot be removed immediately or in isolation, but everyone can be supportive.\footnote{Tech sovereignty is a central topic for the Natives in Tech conference in 2022: \url{https://nativesintech.org/conference/2022}} The creation of a collaborative process is a proposal emerging from the communities themselves: ``Technology as Tequio; technological creation and innovation as a common good'' \cite{aguilar2020}. However, it is not possible to build contemporary data-driven NLP technologies without data. And this opens the discussion regarding Data Sovereignty. First, it is important to mention that the communities have the right to self-determination, and this includes the data that they create. Applying this sovereignty to data refers to having control over the data, knowledge\footnote{See \url{https://indigenousinnovate.org/downloads/indigenous-knowledges-and-data- governance-protocol_may-2021.pdf}} and cultural expressions that are created by these communities. As discussed in this paper, it is important to reach agreements with the communities through consultations and direct collaborations. This includes the licensing and ownership of the final data products. 

\paragraph{Our Findings and Previous Work} Finally, we want to relate our findings to similar discussions in prior work. Most previous concerns and suggestions related to including and consulting people from the communities \cite{bird-2020-decolonising,liu2022not} are aligned with the wishes and desires of the participants in our study. The inclusion of community members as co-authors \cite{liu2022not} should not be an artificial mechanic but more a broad inclusion process, including data and technology sovereignty. This is also aligned with the community building aimed at by \citet{zhang2022can}. Additionally, we should consider that there might exist problematic topics and not underestimate the importance of high-quality translations.

\section{Conclusion}
In this work, which is focused on ethical challenges for MT of Indigenous languages,
we first provided an overview of relevant ethical approaches, ethical challenges for translation in general, and more specific challenges for MT. Afterward, we conducted a case study, for which we interviewed $22$ Indigenous language activists, language teachers, and community leaders from the Americas.  
Our findings aligned with previous findings regarding the need for inclusion and consultation with communities when working with language data. Additionally, our participants expressed a surprisingly strong interest in having MT systems for their languages but also concerns regarding commercial usage, cultural and religious misuse, data, and technological sovereignty. We ended with specific recommendations for the NLP and MT communities, but even more important, an open discussion framework for the indigenous communities. 

\section*{Acknowledgments}
We want to thank all community members, linguists, and language activists who participants in our study. We will also thank the reviewers for their valuable comments and Heriberto Avelino for his useful insights. This project has benefited from
financial support to Manuel Mager by a DAAD
Doctoral Research Grant.

\section*{Limitations}

This study is restricted to the Americas. Therefore the results from this paper can not be generalized, as different indigenous communities or nations might have different pasts. Also, all opinions expressed by the interviewed people are exclusively personal and in should not be interpreted as the general stand of the communities. 
As discussed in the paper, the main aim of this work is not to provide a normative for MT researchers. We rather provide a set of questions and open topics that should be considered when performing MT work with indigenous languages. Nevertheless, we also provide general and broad non-normative recommendations that should be carefully applied to the concrete case of each community.

\section*{Ethical statement}

To ensure the ethics of this work, we followed well-recognized ethical codes: The Australian Institute of Aboriginal and Torres Strait Islander Studies ( AIATSIS)\footnote{\url{https://aiatsis.gov.au/sites/default/files/2020-10/aiatsis-code-ethics.pdf}} and the DOBES code of conduct\footnote{\url{https://dobes.mpi.nl/ethical_legal_aspects/DOBES-coc-v2.pdf}}. As a result, all participants were well informed about the intent of this work, our aims, and the complete anonymization of their answers. Moreover, this work was done with indigenous leadership (as suggested by AIATSIS).

Here we list the ethical issues we found while working on this work and how we try to minimize their impact. First, we were concerned with the data protection of the participants in this study. As for this study, no personal data is required. Therefore, we decided to remove any questions containing any information that could reveal the identity of the participants. Second, as our study aims to get substantial input from the communities, we decided to leave as many open questions as possible and consider the available comments section of each question. All participants were informed about the goals of this project and participated in a free and informed way. To give proper recognition to the participants of this study, we offer an option to be included in the acknowledgment section.

\bibliography{anthology,custom}

\bibliographystyle{acl_natbib}

\appendix

\section{Questionnaire}
\label{appx:quest}
The following questions, together with their answers are listed in this section. 
\begin{enumerate}
\item \textbf{How do you consider yourself}
\begin{enumerate}
    \item Leader of your town or community
    \item Language activist
    \item None of the above
    \item Other, explain
\end{enumerate}

\item \textbf{To what degree do you speak your mother tongue (indigenous / native / native)}
\begin{enumerate}
    \item Perfectly
    \item Fairly good, but with shortcomings
    \item I speak with difficulties
    \item I don't speak it but I do understand
    \item I am not a speaker of the language, but I know some words
\end{enumerate}

\item \textbf{What people / tribe / nation do you belong to?} 
\begin{enumerate}
    \item Open Question
\end{enumerate}

\item \textbf{In your town / nation / tribe there is some restriction to share your language with people outside your community}
\begin{enumerate}
    \item Yes
    \item No
    \item It depends
\end{enumerate}
Comment the question
 
\item \textbf{On a scale from 1-5, how dangerous do you think an MT system would be?}
\begin{itemize}
    \item 1 - Not dangerous
    \item 5 - Dangerous
\end{itemize}
Comment the question

\item \textbf{What advantages would you see with an automatic translation system}
\begin{itemize}
\item Open ended
\end{itemize}
Comment the question

\item \textbf{What dangers would you see in a machine translation system:}
\begin{itemize}
\item Open ended
\end{itemize}
Comment the question

\item \textbf{What topics would you see as positive in which automatic translation can be carried out (there may be several options)}
\begin{enumerate}
\item Medicine and health
\item Medicinal plants
\item Religion and sacred songs
\item Tales and literature and non-sacred songs
\item Laws
\item Commercial matters
\item Talk everyday
\item Science and education
\item Culture and traditions
\item others
\end{enumerate}

\item \textbf{What would you see as damaging topics that should not be machine translated?}
\begin{itemize}
\item Open question
\end{itemize}

\item \textbf{What would you think of a system with poor translation quality?}
\begin{itemize}
\item Not to be used or made available to the public
\item It would be good, to see the progress and be able to help correct it
\end{itemize}

\item \textbf{To create translation systems we need data (text in the native language aligned with text in the foreign language). These data should:}
\begin{enumerate}
\item Be public and without restrictions
\item Be owned by an organ of the community or town.
\item Be the property of the research group that compiled it
\item Be the property of the speakers who made the translation
\end{enumerate}

\item \textbf{In a machine translation project}
\begin{enumerate}
\item Any team of researchers in the world can produce systems 
\item It is preferable to integrate community members
\item Only community members should work on it
\item Only with express permission  of the community body 
\item These investigations should not be carried out
\end{enumerate}

\end{enumerate}

\section{Complete answers of the open questions}
\label{appx:results_open}
All questions refer to the ones enumerated in appendix \ref{appx:quest}.
In table \ref{tab:question6} we present the complete results for the open question 6. 
In table \ref{tab:question7} we present the complete results for the open question 7. 
In table \ref{tab:question9} we present the complete results for the open question 9. 

\begin{table*}[]
    \centering
    \small
\begin{tabular}{l | p{0.5\linewidth} | p{0.5\linewidth}}
    & Original & English Translation \\\hline
    & \bf Qué ventajas tendría un sistema de traducción automática para su idioma? & 
   \bf What dangers would you see in a machine
  translation system (for indigenous languages): \\\hline
1 & Sería practico pero confuso, por las variantes (aprox. mas  de  30 variantes) & It would be practical but confusing, because of the variants (approx. more than 30 variants) \\
2 & Facilitaría el aprendizaje para los ajenos a la lengua y entender cada uno de los nativos la forma estructura de su escritura. & It would facilitate learning for those foreign to the language and for each of the natives to understand the structure of their writing.\\
3 & Fácil acceso & Easy access\\
4 & La unificacion de la lengua& The unification of the language\\
5 & La ventaja de que las personas que no lo hablan, puedan comunicarse con la gente monolingüe & The advantage that people who do not speak it can communicate with monolingual people\\
6 & Ninguna.& None\\
7 & La dispersión del aprendizaje de este & The dispersion of the learning of this \\
8 & Tal vez un aporte de modernidad a la comunidad, preservacion de la lengua nativa. & Perhaps a contribution of modernity to the community, preservation of the native language. \\
9 & Con respeto a los derechos lingüísticos su difusión pertinencia & With respect to linguistic rights its dissemination relevance\\
10 & Facilitaría la enseñanza porque se tendría muchas herramientas de apoyo. & It would facilitate teaching because it would have many support tools. \\
11 & Conocer una aproximación de los significados & Know an approximation of the meanings\\
12 & Ayudaría a las personas que no hablan o ya no hablan el idioma Popti', ayudaría a los niños en las escuelas para aprender el idioma.& It would help people who don't speak or no longer speak the Popti' language, it would help children in schools to learn the language.\\
13 & Contribuiría al uso social de nuestro idioma& It would contribute to the social use of our language\\
14 & El uso de traductores automáticos en espacios como hospitales, oficinas de gobierno, etc. & The use of automatic translators in spaces such as hospitals, government offices, etc.\\
15 & Contribuiría al Status de las lenguas indígenas y quizá también Corpus.& It would contribute to the Status of indigenous languages and perhaps also Corpus.\\
16 & Solo seria útil para conocer vocabulario y pequeñas frases quizás. & It would only be useful to learn vocabulary and small phrases perhaps.\\
17 & hacer la lengua mas visible y más accesible a personas fuera de la comunidad de habla& make the language more visible and more accessible to people outside the speaking community\\
18 & Mantener la lengua a través del tiempo y la distancia & hold the tongue through time and distance\\
19 & Mucha ventaja para las personas que han estado interesados en aprender este idioma.& Much advantage for people who have been interested in learning this language.\\
20 & nos ayudaría a normalizar la presencia del kichwa en un medio actual de comunicación.& It would help us to normalize the presence of Kichwa in a current means of communication. \\
21 & These are great for interview videos or audio. I don't see anything wrong with this.& \\
22 & Ayudar aprender nuestra lengua& Help learn our language\\
      
\end{tabular}
    \caption{Answers for the open question 6 (see \S\ref{appx:quest}). All translations are automatic translations with human editing. If the original is in English, we leave the translation field blank.}
    \label{tab:question7}
\end{table*}

\begin{table*}[]
    \centering
    \small
\begin{tabular}{l | p{0.5\linewidth} | p{0.5\linewidth}}
    & Original & English Translation \\\hline
    & \bf Podría usted commentar sobre algún posible problema que podría generar un sistema de traducción automática para su idioma?  & 
   \bf What dangers would you see in a machine
translation system \\\hline
1 & Por la variedad de la Lengua, algunas palabras significan diferente, no es recomendable un diccionario estandar para la lengua, si por variante. & Due to the variety of the language, some words mean differently, a standard dictionary for the language is not recommended, if by variant.\\
2 &  Los conceptos culturales.  En su defecto hay conceptos lingüísticos culturales que solo se entienden en la lengua nativa.  De igual forma, encontraremos conceptos ajenos a nuestra lengua que no son traducibles. & Cultural concepts. Failing that, there are cultural linguistic concepts that are only understood in the native language. In the same way, we will find concepts foreign to our language that are not translatable. \\
3 &  Falta de recursos de información  & Lack of information resources\\
4 &  La gente de cada región  se aferra a su propia variante  & The people of each region cling to their own variant \\
5 &  No se puede entender estrictamente como peligroso, pero si un poco difícil porque existen muchas variantes & It cannot be strictly understood as dangerous, but it is a bit difficult because there are many variants\\
6 &  La desnaturalización del idioma. Al tratarse de un idioma donde no imperan los sustantivos, se carece de artículos y de algunas preposiciones, además de ser aglutinante, el trabajo es mucho más complicado que en otros idiomas. Además, la existencia de tantas variantes haría que el proyecto fuera poco o nada rentable y llevaría a los "expertos" a un intento por homologar el habla, lo cual sería un error tremendo. & The denaturation of the language. Being a language where nouns do not prevail, articles and some prepositions are lacking, in addition to being agglutinative, the work is much more complicated than in other languages. In addition, the existence of so many variants would make the project unprofitable or unprofitable and would lead the "experts" to attempt to standardize speech, which would be a tremendous mistake.\\
7 &  Ayudaría a el aprendizaje de la lengua. & It would help to learn the language.\\
8 &  Ninguna, de hecho tenemos avances y una comunidad internacional(Peru, Bolivia, Argentina y Chile) de aymara formada con este objetivo. & None, in fact we have progress and an international community (Peru, Bolivia, Argentina and Chile) of Aymara formed with this objective.\\
9 &  Su comercialización  & the commercialization\\
10 & Hay algunas palabras que no se pueden traducir, porque al hacerlo pierde el sentido, además de que esta lengua tiene otras grafías que están ausentes en el sistema consonántico y vocálico del español (que es como la referencia de todo lo que se hace). Además de que la lengua O’dam es aglutinante, las correspondencias entre palabras u oraciones no siempre será uno a uno porque las palabras en O’dam son largas y puede significar toda una frase.  Es una lengua aglutinante pues. & There are some words that cannot be translated, because by doing so they lose their meaning, in addition to the fact that this language has other spellings that are absent in the consonantal and vowel system of Spanish (which is like the reference for everything that is done). In addition to the fact that the O'dam language is agglutinative, the correspondences between words or sentences will not always be one to one because the words in O'dam are long and can mean a whole sentence. It is an agglutinative language.\\
11 & Tomar en cuenta la cultura, existen elementos culturales que deben tomarse en cuenta  & Take culture into account, there are cultural elements that must be taken into account\\
12 & Ninguno. & None\\
13 & Que no sea pertinente cultural y lingüísticamente  & Not culturally and linguistically adequate\\
14 & Que por "practicidad" se priorice el uso de traductores automáticos en lugar de aprender la lengua.  & That for "practicality" the use of automatic translators is prioritized instead of learning the language. \\
15 & Podrían sin lugar a dudas desvirtuar el adecuado uso del idioma. El quechua es un idioma contextual. & They could undoubtedly distort the proper use of the language. Quechua is a contextual language.\\
16 & la pregunta anterior es muy ambigua, peligroso en qué sentido, para sus hablantes? para el linguista? ... un problema que podría generar es que no toda la comunidad de habla esté de acuerdo en poner la langua al alcance de todos & the previous question is very ambiguous, dangerous in what sense, for its speakers? for the linguist? ... a problem that could arise is that not the entire speech community agrees to make the language available to everyone\\
17 & Lo primero es que no hay un teclado especial  & The first thing is that there is no special keyboard\\
18 & Sería un poco difícil de realizar un translate automático del chol, ya que es un lengua compleja y a veces solamente se entiende viendo, estando o conociendo el contexto del habla. &  It would be a bit difficult to translate Chol automatically, since it is a complex language and sometimes it is only understood by seeing, being or knowing the context of the speech.\\
19 & ninguno & none\\
20 & siempre y cuando la comunidad este involucrada. & as long as the community is involved. \\
21 & As long as the community owns the language corpus then I think it would be helpful. I have concerns of other entities, academic or commercial, being involved in handling the corpus. & \\
22 & Entes externos podrían aprovecharse económicamente & External entities could take advantage economically\\
  
\end{tabular}
    \caption{Answers for the open question 7 (see \S\ref{appx:quest}). All translations are automatic translations  with human editing. If the original is in English, we leave the translation field blank.}
    \label{tab:question9}
\end{table*}

\begin{table*}[]
    \centering
    \small
\begin{tabular}{l | p{0.5\linewidth} | p{0.5\linewidth}}
    & Original & English Translation \\\hline
    & \bf Qué temas vería usted cómo perjudiciales en los que no se debería realizar traducción automática para su lengua? & 
   \bf What would you see as damaging topics that should not be machine translated? \\\hline
1 &  Ninguno & None\\
2 &  Leyes, medicina y salud, ciencia, cuestiones mercantiles, religión y canciones sagradas,  & Law, medicine and health, science, business matters, religion and sacred songs,\\
3 &  Derecho  & Laws\\
4 &  Las groserías & the bad words\\  
5 &  No veo problema alguno en la traducción de todos lados temas & I do not see any problem in the translation of all sides topics\\
6 &  ninguno & None\\
7 &  Ninguno. & None\\
8 &  Ninguna, pero hay muchas versiones de aymara y con influyencia de castellano. & None, but there are many versions of Aymara and with Castilian influence.\\
9 &  Temas que atenten contra la vida orgánica  & Issues that threaten organic life\\
10 & Lo sagrado ( discursos rituales, cantos, etc.) por eso de que las palabras en el caso del Mucha fuerza al pronunciarse razón por la que no deben decirse fuera de su contexto.  & The sacred (ritual speeches, songs, etc.) That is why the words in the case of Much force when pronounced, reason why they should not be said out of context.\\
11 & Significados culturales & cultural meanings\\ 
12 & Religión occidenta & western religion\\
13 & Religión & Religion\\
14 & Conocimientos considerados sagrados.  & Knowledge considered sacred.\\
15 & Las traducciones automáticas no son completamente válidas ni aún en los idiomas modernos. & Machine translations are not completely valid even in modern languages. \\
16 & ningun tema & no topic\\
17 & Situaciones políticas y religiones a menos que sea del interés de la persona  & Political situations and religions unless it is in the interest of the person\\
18 & Los cantos sagrados, como los de un curandero. & The sacred songs, like those of a healer.\\
19 & ninguno & none\\
21 & Medicina y Salud, Plantas medicinales, Religión y canciones sagradas & Medicine and Health, Medicinal plants, Religion and sacred songs \\
20 & Anything ceremonial & \\
22 & curaciones y cuestiones personales & cures and personal issues
\end{tabular}
    \caption{Answers for the open question 9 (see \S\ref{appx:quest}). All translations are automatic translations with human editing. If the original is in English, we leave the translation field blank.}
    \label{tab:question6}
\end{table*}

\newpage
\section{Spanish Version: \textit{Consideraciones éticas para la traducción automática de lenguas indígenas: dar voz a los hablantes}}
\label{sec:spanish}

\textit{Incluimos una traducción del texto original en inglés, para facilitar la discusión del mismo en los países de habla hispana del continente Americano. A si mismo esta traducción nos permite compartir los resultados de nuestra investigación con los participantes de las entrevistas y encuestas. La traducción fue realizda con ayuda de un sistema de tradicción automática (Google Translate) y mejorada con edición manual. Todas las referencias son enlazadas al original en inglés.}

\paragraph{Resúmen} En los últimos años, la traducción automática se ha vuelto muy exitosa para los pares de idiomas con altos recursos. Esto también ha despertado un nuevo interés en la investigación sobre la traducción automática de idiomas de bajos recursos, incluidos los idiomas indígenas.
Sin embargo, estos últimos están profundamente relacionados con los grupos indígenas que los hablan (o solían hablar). La recopilación de datos,
modelar y desplegar sistemas de traducción automática da como resultado nuevas cuestiones éticas que deben abordarse.
Motivados por esto, primero examinamos la literatura existente sobre las consideraciones éticas para la documentación, la traducción y el procesamiento general del lenguaje natural para las lenguas indígenas. Posteriormente, realizamos y analizamos un estudio de entrevistas para arrojar luz sobre las posiciones de los líderes comunitarios, maestros y activistas lingüísticos con respecto a las preocupaciones éticas sobre la traducción automática de sus idiomas. Nuestros resultados muestran que la inclusión, en diferentes grados, de hablantes nativos y miembros de la comunidad es vital para realizar una mejor investigación y más ética sobre las lenguas indígenas.

\subsection{Introducción}

Con el avance de los sistemas de traducción automática (MT) basados en datos, se ha vuelto posible, con diversos grados de calidad, traducir entre cualquier par de idiomas. La única condición previa es la disponibilidad de suficientes datos monolingües \cite{lample2018unsupervised,artetxe2018unsupervised} o paralelos \cite{vaswani2017attention,bahdanau2015neural}. Hay muchas ventajas de tener sistemas de traducción automática de alto rendimiento. Por ejemplo, aumentan el acceso a la información para los hablantes de idiomas indígenas \cite{mager-etal-2018-challenges} y pueden ayudar en los esfuerzos de revitalización de estos idiomas \cite{zhang2022can}.

La investigación sobre la traducción automática, así como el procesamiento del lenguaje natural (NLP), en general, se está moviendo hacia entornos de bajos recursos y modelos multilingües. Por lo tanto, la comunidad de NLP necesita abrir la discusión sobre las repercusiones y las mejores prácticas para la investigación en lenguas indígenas (que en la mayoría de los casos también son de bajos recursos) ya que las lenguas naturales no pueden existir sin una comunidad de personas que usan (o han usado tradicionalmente).

Los idiomas indígenas difieren además de los más utilizados de una manera crucial: son comúnmente hablados por pequeñas comunidades, y muchas comunidades usan su idioma (además de otras características) como un delimitador para definir su propia identidad \cite{palacios2008lengua,enriquez2019rol}, y tienen en muchos casos también un cierto grado de peligrosidad. Además, en algunos casos, información altamente confidencial, como aspectos secretos de su religión, ha sido codificada con la ayuda de su idioma \cite{carlos2016richard}. Esta es la razón por la que, en los últimos años, se han iniciado debates sobre enfoques éticos para estudiar lenguas en peligro \cite{smith2021decolonizing,liu2022not}. Cuando consideramos el pasado (y presente) de algunas de las comunidades que hablan estos idiomas, encontraremos una historia colonial, donde la investigación no es la excepción \cite{bird-2020-decolonising}.
Por lo tanto, es posible traspasar los límites éticos cuando se utilizan metodologías típicas de recopilación de datos y NLP \cite{dwyer2006ethics}.

En este trabajo, exploramos los conceptos básicos de ética relacionados con la MT de lenguas en peligro de extinción con un enfoque especial en las comunidades indígenas, examinando trabajos previos sobre el tema. Para comprender mejor las expectativas y preocupaciones relacionadas con el desarrollo de sistemas de MT para las comunidades indígenas, luego realizamos un estudio de entrevistas con 22 activistas lingüísticos, profesores de idiomas y líderes comunitarios que son miembros de comunidades indígenas de América. Además, también realizamos diálogos 1:1 con dos participantes del estudio para profundizar nuestra comprensión del asunto. El objetivo es responder las siguientes preguntas de investigación: \textit{¿Cómo quieren participar los miembros de la comunidad en el proceso de traducción automática y por qué?} \textit{¿Hay temas sensibles que no son éticos para traducir, modelar o recopilar datos sin ¿el permiso explícito de la comunidad?} \textit{¿Cómo podemos recopilar datos de manera ética?}

Sorprendentemente, la mayoría de los participantes de la encuesta ven positivamente la traducción automática de sus idiomas. Sin embargo, creen que la investigación sobre sus idiomas debe realizarse en estrecha colaboración con los miembros de la comunidad. El acceso abierto a los descubrimientos y recursos de investigación también se valora mucho, así como la alta calidad de las traducciones resultantes. Las entrevistas personales también lo confirmaron.
Por lo tanto, nuestro hallazgo más importante es que es fundamental trabajar en estrecha colaboración con las comunidades para comprender temas éticos delicados al desarrollar sistemas de traducción automática para lenguas en peligro de extinción.

\subsection{Definiendo ``lenguaje en peligro''}

Los términos que se usan con frecuencia en la NLP son \textit{lenguaje de bajos recursos}, \textit{lenguaje de pocos recursos} y \textit{configuración de bajos recursos}. Esos términos no resaltan el hecho de que muchos idiomas de bajos recursos también están en peligro \cite{liu2022not}. En cambio, enfatizan el problema crítico del aprendizaje automático de lograr que un enfoque basado en datos funcione bien con una cantidad de datos disponible menor a la ideal (o simplemente menos datos que los que se han usado para otros idiomas). En este caso, se necesitan innovaciones algorítmicas o tecnológicas para cerrar la brecha de rendimiento entre los lenguajes de muchos recursos y los de pocos recursos. Esto implica además que tener pocos recursos no es una propiedad de un idioma sino un término que solo tiene sentido en el contexto de una tarea o tareas en particular.

En contraste, el término \textit{lenguaje en peligro} se refiere a un idioma con cierto grado de peligro para su existencia.\footnote{En este documento, discutiremos solo los idiomas creados no artificialmente.}
  Los idiomas en peligro son relevantes para nuestro estudio, ya que la mayoría de los idiomas indígenas también están en peligro \cite{hale1992endangered}.
Según la clasificación de la UNESCO, los idiomas en peligro \cite{moseley2010atlas} se pueden clasificar en las siguientes categorías diferentes:

\begin{itemize}
     \item \textit{saludables}: hablado por todas las generaciones;
     \item \textit{vulnerable}: restringido solo a un determinado dominio (es decir, dentro de la familia);
     \item \textit{definitivamente en peligro}: no tiene niños que hablen el idioma;
     \item \textit{en grave peligro}: sólo lo hablan las personas mayores;
     \item \textit{peligro crítico}: sólo quedan hablantes con un conocimiento parcial, y lo usan con poca frecuencia;
     \item \textit{extinto}, cuando ya no haya personas capaces de hablar el idioma.
\end{itemize}

Los idiomas pueden estar en peligro debido a razones sociales, culturales y políticas; más comúnmente conquistas y guerras, presiones económicas, políticas lingüísticas de los poderes políticos, asimilación de la cultura dominante, discriminación y estandarización lingüística \cite{austin2013endangered}. Como podemos ver, el problema de cómo una lengua se pone en peligro involucra factores que deben ser abordados en el enfoque ético de cualquier estudio.
Por el lado del aprendizaje automático, surge un desafío adicional: los datos de los idiomas en peligro no están fácilmente disponibles (o, de hecho, no están disponibles en absoluto), ya que estos idiomas tienen una producción de medios limitada \citep[programas de televisión, literatura, blogs de Internet; ][]{hamalainen2021endangered}. Una posible fuente de datos para estos idiomas son los documentos ya existentes en forma de libros, registros y archivos \cite{bustamante-etal-2020-data}.

\subsection{Ética y MT}
\paragraph{Ética y datos}

El estudio de las lenguas en peligro en las comunidades indígenas tiene una larga historia, y las preguntas más destacadas se centran principalmente en el desafío de la recopilación de datos \cite{smith2021decolonizing}.

Una de las formas comunes de esto es usar la ética normativa (deontología). Ejemplos de pautas relevantes incluyen las del Instituto Australiano de Estudios de los Aborígenes e Isleños del Estrecho de Torres;\footnote{\url{https://www.jstor.org/stable/pdf/26479543.pdf}} la Declaración ética de la Sociedad Lingüística de América;\footnote{\url{https://www.linguisticsociety.org/content/lsa-revised-ethics-statement} \url{-approved-july-2019}} y el código de conducta DOBES.\footnote{ \url{https://dobes.mpi.nl/ethical_legal_aspects/DOBES-coc-v2.pdf}} Estas listas son el resultado de amplias discusiones que han tenido lugar durante décadas. En este debate también, las voces indígenas se alzaron dentro de la academia \cite{smith2021decolonizing}.

Pero, ¿por qué tenemos tantos intentos de establecer un código ético para el trabajo de campo lingüístico? Cuando se trata de trabajar con sociedades humanas, no existen soluciones fáciles para los dilemas éticos que surgen \cite{dwyer2006ethics}. Cada situación requiere un trato y compromiso único. Por eso, además de la creación de un marco lo más general posible, la aplicación concreta de tales principios implica una discusión continua.
\newcite{dwyer2006ethics} sugiere documentar las cuestiones y preocupaciones éticas que surgen durante el curso de un proyecto de investigación y la forma en que se abordan estas cuestiones, de modo que otros investigadores puedan aprender de la experiencia. Si bien un código de conducta o principios es bueno, corre el riesgo de introducir regulaciones demasiado complicadas, o incluso inadecuadas, relegando esta necesaria discusión.

En general, podemos resumir los principios que aparecen en todas las listas sugeridas bajo tres grandes temas:
\begin{itemize}
     \item \textit{Consulta, Negociación y Entendimiento Mutuo}. El derecho a la consulta de los pueblos indígenas está estipulado en el convenio 167 de la Organización Internacional del Trabajo \cite{ilo1989c169} y establece que ``tienen derecho a preservar y desarrollar sus propias instituciones, idiomas y culturas''. Por lo tanto, informar a la comunidad sobre la investigación planificada, negociar un posible resultado y llegar a un acuerdo mutuo sobre las direcciones y los detalles del proyecto debe ocurrir en todos los casos.
     \item \textit{Respeto a la cultura local y participación}. Como cada comunidad tiene su propia cultura y visión del mundo, los investigadores, así como cualquier organización gubernamental interesada en el proyecto, deben estar familiarizados con la historia y las tradiciones de la comunidad. Además, se debe recomendar que investigadores locales, oradores o gobiernos internos se involucren en el proyecto.
     \item \textit{Compartir y distribuir datos e investigaciones}. El producto de la investigación debe estar disponible para su uso por parte de la comunidad, para que puedan aprovechar los materiales generados, como documentos, libros o datos.
\end{itemize}
    
Algunos de estos principios comúnmente acordados están abiertos a un acuerdo
deben adaptarse a situaciones concretas, lo que podría no ser fácil de hacer a través de un enfoque general. Por ejemplo, el proceso de documentación creará datos, y la propiedad de estos datos es una fuente importante de discusión (cf. Secciones \ref{sec:study}, \ref{sec:discussion}). Aquí, los puntos de vista tradicionales de las comunidades pueden contradecir el sistema jurídico de un país \cite{daes1993discrimination}. Este problema no tiene una solución simple y debe considerarse cuidadosamente al recopilar datos.

Un llamado adicional de estas fuentes es descolonizar la investigación y dejar de ver a las comunidades indígenas como fuentes de datos, sino como personas con su propia historia
\cite{smith2021decolonizing}. El divorcio actual entre los investigadores y las unidades culturales de las comunidades puede llevar a reforzar el legado colonial \cite{leonard2020producing}.

Como comentario final, queremos discutir la suposición común de que cualquier discusión ética debe terminar con una configuración normativa para un campo. Esto reduciría la trducción de toma de decisiones colectivas de los pueblos indígenas a normas que permitan un abordaje individual del asunto \cite{meza2017etica}. Esto tampoco permitiría comprender las cuestiones éticas con su propia cosmovisión indígena comunal \cite{salcedo2016vivir}. Por lo tanto, en este texto pretendemos abrir el debate ético de la MT a los investigadores de la NLP ya las comunidades indígenas a partir de la inclusión y el diálogo.

\paragraph{Ética y \textit{Traducción} humana}

Para una traducción exitosa, la inclusión de todos los participantes es importante, requiriendo su participación equitativa, informal y orientada a la comprensión \cite{nissing2009grundpositionen}.
Para \newcite{rachels1986elements}, la concepción mínima de la moralidad es que cuando damos "igual peso a los intereses de cada individuo afectado por la decisión de uno". La pregunta es cómo las intenciones de los autores se relacionan con la otredad de la cultura de origen, con su valores culturalmente específicos \cite{chesterman2001proposal}.
Según \newcite{doherty2016translations}, ``los estudios del proceso de traducción surgieron para centrarse en el traductor y el proceso de traducción en lugar del producto final'', incorporando diseños de métodos mixtos para obtener observaciones objetivas.
Un ejemplo bien documentado del mal uso no ético de la traducción es la aplicación de la traducción como instrumento para la dominación colonial. El principal objetivo de esta visión colonialista era ``civilizar a los salvajes'' \cite{ludescher2001instituciones}. Por ejemplo, el instituto lingüístico de verano (ILV)\footnote{ILV se describe a sí mismo como ``.. una organización mundial sin fines de lucro basada en la fe que trabaja con comunidades locales de todo el mundo para desarrollar soluciones lingüísticas que amplíen las posibilidades de una vida mejor. . Las principales áreas de contribución de SIL son la traducción de la Biblia, la alfabetización, la educación, el desarrollo, la investigación lingüística y las herramientas lingüísticas.''. \url{https://www.sil.org/}} se utilizó para este objetivo durante el siglo XX en países con culturas indígenas, traduciendo la Biblia y tratando de provocar un cambio cultural\footnote{El papel del SIL es controvertido, y no se puede resumir en una sola declaración. En nuestro enfoque, solo nos referimos al papel jugado en relación con el cambio cultural. En muchos casos, las comunidades que hicieron traducir los textos religiosos ya eran cristianas, dadas las acciones de colonización previas. Sin embargo, también hay casos en los que comunidades no cristianas tuvieron Biblias y otros textos religiosos traducdios a su idioma, con fines misioneros. Esto desencadenó divisiones comunitarias. Por ejemplo, la traducción de los textos religiosos al Wixarika \cite{fernandez2022libertad}. Esto también sucedió en la Comunidad de Zoquipan (en el estado mexicano de Nayarit), donde los cristianos, utilizando la Biblia traducida del ILV, desencadenaron un conflicto interno en la comunidad (el primer autor es parte de esta comunidad). Para el lector interesado, también recomendamos \newcite{dobrin2009sil} artículo introductorio.} en estas comunidades \cite{delvalls1978instituto,errington2001colonial,carey2010lancelot}. Por supuesto, estas prácticas no son nuevas y se pueden encontrar a lo largo de la historia \cite{gilmour2007missionaries}. Es esencial tener en cuenta que la investigación no ética aún puede brindar material y conocimientos útiles, por ejemplo, para la revitalización del idioma \cite{premsrirat2003language}, pero podría causar daño a la comunidad objetivo.

\paragraph{Ética y \textit{Traducción} automática}

En el contexto de la investigación de NLP, los hablantes no están directamente involucrados cuando se entrena un modelo \citep{pavlick2014language}. En contraste, los procesos de recolección de datos \cite{fort2011crowdsourcing} y evaluación humana \cite{couillault-etal-2014-evaluating} interactúan directamente con los hablantes y, por lo tanto, tienen una importancia central en relación con la ética. Esto también es válido para el servicio de traducción final, que interactuará con el público en general.

La recopilación de datos es el primer y más evidente problema cuando se trata de traducción. Los sistemas MT neuronales modernos requieren una gran cantidad de datos paralelos para entrenarse de manera óptima \cite{junczys2019microsoft}. Una forma de obtener datos es mediante el crowdsourcing \cite{fort2011crowdsourcing}. Sin embargo, este tipo de trabajo puede estar mal pagado y puede constituir un problema para las condiciones de vida de los trabajadores \cite{schmidt2013good}.
Además, la privacidad de los datos no es trivial de manejar. Los sistemas deben ser capaces de filtrar información confidencial.

El problema de los sesgos de codificación\footnote{También es importante tener en cuenta las características tipológicas que podrían dificultar esto. Un ejemplo son los lenguajes polisintéticos y los lenguajes sin codificación de género \cite{klavans-2018-computational}.}, como el sesgo de género \cite{stanovsky2019evaluating}, también es una preocupación ética para MT.
También es necesario revelar las limitaciones y problemas con ciertos sistemas \cite{leidner-plachouras-2017-ethical}.

La investigación de la NLP también se puede utilizar como un instrumento político de poder, donde podemos observar las relaciones mutuas entre el lenguaje, la sociedad y el individuo que "también son la fuente de los factores de impacto social de la NLP" \cite{horvath2017language}. De esta forma, la traducción automatica se puede aplicar como un instrumento para cambiar la cultura de las minorías como en la traducción tradicional (cf. Sección \ref{subsec:ethics_translation}). Entonces, los colonizadores usaron la traducción como medio de control imperial y expropiación \cite{cheyfitz1997poetics,niranjana10siting}. La asimetría de poder es la causa de la dominación, donde las culturas subalternas que se inundan con ``imposiciones de materiales e idiomas extranjeros'' son un peligro real para las culturas minoritarias \cite{tymoczko2006translation}.
\newcite{schwartz2022primum} discute la necesidad de descolonizar el enfoque científico de la comunidad de NLP en su conjunto, expresando la necesidad de que los investigadores sean conscientes de la historia y los aspectos culturales de las comunidades que utilizan los idiomas con los que están trabajando. Además, propone que nuestra investigación debe tener una obligación de brindar algún beneficio de nuestros estudios a las comunidades, una obligación de rendición de cuentas (y por lo tanto estar en contacto directo con sus organismos rectores) y una obligación de no tener intenciones que contravengan los intereses de las comunidades. El hecho de que muchos sistemas de traducción hoy en día sean multilingües\footnote{Los sistemas multilingües se refieren en NLP a sistemas capaces de traducir un conjunto de idiomas desde y hacia el inglés. En algunos casos, también son capaces de traducir entre idiomas donde el inglés no está involucrado.} también resultan en más desafíos multiculturales \cite{hershcovich2022challenges}.

Finalmente, también queremos resaltar la importancia de discutir los sistemas MT en una configuración de texto a texto. El uso del texto está limitado a ciertos temas y varía de una comunidad a otra. Por ejemplo, el wixarika y el quechua, idiomas que se hablan en todas las generaciones, se usan de manera escrita principalmente en aplicaciones de mensajería privada (como WhatsApp), pero también tienen una prolífica generación de publicaciones en Meme y Facebook\footnote{Por ejemplo, los memes wixarika: \ url{https://www.facebook.com/memeswixarika2019}, grupo de habla quechua: \url{https://www.facebook.com/groups/711230846397383/}}. Incluso si una determinada comunidad no adopta ampliamente la tradición escrita, existen, como mínimo, obligaciones legales de los Estados hacia las lenguas indígenas. Por ejemplo, algunas constituciones reconocen las lenguas indígenas como lenguas nacionales (es decir, México y Bolivia), obligando al Estado a la responsabilidad de traducir todas las páginas oficiales, documentos, leyes, etc., a las lenguas indígenas. Esto no se ha implementado, y este caso es un caso de aplicación muy valioso para la traducción automática para ayudar a la traducción humana. Sin embargo, nuestros hallazgos también se aplican a la traducción de voz a texto y a las tareas de voz a voz que cubrirían todos los idiomas, incluso sin tradición escrita.

\subsection{Las opiniones de los oradores}

Es importante incluir la opinión y la visión de los hablantes de lenguas en peligro de extinción en la investigación de la NLP, especialmente para temas como la MT. Por lo tanto, llevamos a cabo un estudio de encuesta con 22 activistas lingüísticos, maestros y líderes comunitarios de  América. Es importante destacar que nuestro objetivo principal no es solo recopilar información cuantitativa sobre las preguntas éticas relacionadas con la traducción automática para sus idiomas, sino también recopilar información cualitativa al pedirles que amplíen sus respuestas. Además, también realizamos una entrevista con un subconjunto de dos participantes del estudio de entrevista inicial.

\subsection{Diseño del estudio}

Enfocamos nuestro estudio en el continente Americano,\footnote{Diferentes partes del mundo tienen niveles muy diferentes de consciencia, no solo por la historia colonial sino precisamente por las interacciones con los trabajadores de campo.} seleccionando las siguientes comunidades: aymara, chatino, maya, mazateco , mixe, nahua, otomí, quechua, tenek, tepehuano, kichwa de Otavalo y zapoteco. Queremos señalar que nuestro estudio no pretende representar una opinión general de todos los pubelos indígenas, ni es una declaración general final sobre el tema. Es un estudio de caso que aflora las opiniones de grupos específicos de hablantes de lenguas indígenas. Además, las opiniones de las personas entrevistadas son propias y
no representan necesariamente la opinión de sus tribus, naciones o comunidades.

\paragraph{Aspectos Cuantitativos y Cualitativos}

Para el estudio cuantitativo se utilizó una encuesta. Las encuestas son una técnica bien establecida para ser utilizada con comunidades indígenas con una extensa historia y son utilizadas y documentadas por clásicos como Edward Tylor, Anthony Wallace, Lewis Henry Morgan. Esto también es cierto para reconocidos antropólogos sociales mexicanos (indígenas comprometidos) \cite{jimenez1985teoria,alfredo1978teoria}.

Para la parte cualitativa, revisamos documentos de posición existentes y artículos de investigadores y activistas indígenas. Además, utilizamos preguntas abiertas en la encuesta, lo que permite extender la visión puramente cuantitativa a una cualitativa. Finalmente, realizamos dos entrevistas 1 a 1 con un activista (mixe) y un lingüista (chatino).

\paragraph{Reclutamiento de participantes} Nos pusimos en contacto con
participantes potenciales de tres maneras. Nuestro primer enfoque fue establecer comunicación a través de los sitios web oficiales del proyecto de los participantes potenciales o cuentas públicas en línea. Esto incluye las páginas de correo electrónico, Twitter, Facebook e Instagram. Nuestro segundo enfoque fue contactar directamente a las personas de nuestro grupo objetivo con quienes al menos uno de los coautores ya ha establecido una relación de trabajo. Por último, también publicamos una convocatoria de participación en las redes sociales,
y comprobamos si los voluntarios pertenecen a nuestro grupo objetivo. Los objetivos de nuestra investigación, así como el alcance y manejo de datos, se explican directamente a cada participante y también se incluyen en el formulario final. No recopilamos ninguna información personal sobre los participantes, como nombre, sexo, edad, etc.
Todos los participantes del estudio son voluntarios.

\paragraph{Cuestionario} Nuestro estudio consta de 12 preguntas. Las primeras tres preguntas son  generales: preguntan por la tribu, nación o pueblo indígena al que pertenece el participante; si se identifica a sí mismo como activista, líder comunitario o maestro, y por su fluidez en su idioma. Las preguntas restantes se enfocan en políticas de datos, políticas de inclusión, beneficios y peligros de los sistemas MT y mejores prácticas de investigación.
El cuestionario completo está disponible en el apéndice A. Las preguntas están disponibles en inglés y español, pero solo se ha completado un formulario en inglés, mientras que el resto se ha completado en español. Por lo tanto, los autores han traducido automáticamente todos los comentarios que se muestran en este documento al Inglés.

\subsection{Resultados}

Los resultados del estudio se pueden ver en la Figura \ref{fig:results}. Además, también discutimos las respuestas abiertas a cada pregunta para brindar más información.

\paragraph{Inclusión de hablantes nativos y permisos para estudiar el idioma} La figura \ref{fig:results}(a) muestra que el 77,3\% de los participantes informan que su comunidad no tiene restricciones con respecto a compartir su idioma con personas externas.
Los comentarios para esta pregunta muestran que muchos participantes están
orgullosos de su idioma y herencia: ``Somos solidarios y compartimos nuestras raíces. Orgullosos de quien nos visita'' Incluso encontramos pronunciamientos más fuertes contra la prohibición de compartir: ``Nadie tiene derecho a restringir la difusión de la lengua''.
Sin embargo, también existen comunidades con restricciones. Por lo tanto, concluimos que los investigadores no pueden asumir por defecto que todos los grupos indígenas estarían de acuerdo en compartir información sobre su idioma o que estarían contentos con la investigación sobre el mismo.

\paragraph{Beneficios y peligros de los sistemas MT}
La figura \ref{fig:results}(b) muestra que una gran mayoría de nuestros participantes piensa que un sistema de traducción automática para su idioma sería beneficioso. Sin embargo, también hay un número importante de personas que ven al menos algún grado de peligro. En este caso, necesitamos mirar los comentarios de los participantes para entender sus preocupaciones. En primer lugar, encontramos que una de las principales preocupaciones de los participantes es la calidad de la traducción.
El miedo a las traducciones inadecuadas de los términos culturales también es importante. En la Tabla \ref{tab:dangers}, podemos ver un conjunto de comentarios que ilustran estos temores. Un comentario interesante se refiere al miedo a la estandarización del lenguaje de los participantes, lo que podría conducir a una pérdida de diversidad. En la misma tabla, también podemos ver los beneficios que esperan los participantes, principalmente en educación y en elevar el estatus y la utilidad de sus idiomas.

La tabla \ref{tab:topics} muestra algunas respuestas a la pregunta abierta sobre posibles temas que podrían causar daño a la comunidad. La mayoría de las respuestas no pudieron identificar ningún tema posible que pudiera ser peligroso. Sin embargo, la segunda respuesta más frecuente estuvo relacionada con la religión. Algunas respuestas están preocupadas de que se puedan revelar antiguos secretos ceremoniales. Otros también muestran preocupación por la influencia de las religiones occidentales. Esto nos lleva a la pregunta de si la Biblia \cite{christodouloupoulos2015massively,mccarthy2020johns,agic2019jw300} es adecuada para usar como nuestro corpus predeterminado para MT, cuando se trata de un idioma indígena. Finalmente, también algunas respuestas expresaron que el uso de lenguas indígenas en la organización interna de la comunidad podría estar en peligro con los sistemas de MT. En contraste, la figura \ref{fig:results}(c) muestra los temas que registraron esa evaluación más positiva: charlas cotidianas (15), ciencia y educación (14), cultura y tradiciones (14), y medicina y salud (14).

\paragraph{Participación de miembros de comunidades indígenas en la investigación} La figura \ref{fig:results}(d) muestra que los participantes de nuestro estudio piensan que es importante incluir personas de las comunidades objetivo en los proyectos de investigación.
Esto confirma la experiencia en lingüística, donde encontraron un patrón similar \cite{smith2021decolonizing} (ver \S\ref{subsec:ethics_documentation}).
Es importante señalar que solo se indicó una respuesta de que se necesita permiso oficial para realizar el estudio.
En los comentarios se mencionó el derecho a la consulta, junto con las ventajas de involucrar a los miembros de la comunidad en la investigación: ``Es preferible [integrar a personas de la comunidad] para obtener un buen sistema, y no solo tener aproximaciones, porque solo los miembros de la cultura saben cómo se usa el idioma.”; ``Para que se enriquezca el vocabulario y no se impongan algunas palabras que no existen.''; ``Realizar actividades donde la comunidad se pueda involucrar, ganar- ganar.''.

\paragraph{Uso de datos y calidad de traducción} Con respecto a la propiedad y accesibilidad de los datos, encontramos diversos conjuntos de respuestas. Primero, la Figura \ref{fig:results}(e) muestra muchas opiniones diferentes. 
En general, podemos decir que existe una fuerte tendencia a favorecer que los datos deben estar disponibles públicamente.
Sin embargo, cuando se trata de la propiedad de los datos, las opiniones son más diversas. Sorprendentemente, un número importante de participantes ($17\%$) piensa que el grupo de investigación externo debería ser el propietario de los datos. Sin embargo, un mayor número de participantes piensa que los datos deberían ser propiedad de la comunidad ($29,4\%$), y un 20,6\% piensa que deberían ser propiedad de los ponentes que participan en la investigación. Este es un tema complejo, ya que las normas tradicionales y los sistemas legales modernos interactúan (cf. Sección \ref{subsec:ethics_documentation}). En los comentarios encontramos tristes ejemplos de desconfianza en las instituciones académicas. Por ejemplo, un comentario habla de los problemas previos de su tribu, ya que las grabaciones y otros materiales tomados por los lingüistas no son accesibles para ellos: ``Cuidado con las instituciones académicas ya que actualmente tenemos problemas para acceder a las grabaciones que pertenecen a académicos y bibliotecas y no son públicas.”. Sin embargo, en general, vemos una amplia gama de opiniones: ``Debe valorarse el trabajo de los pocos que se toman en serio la identidad lingüística'',
``Puede ser público pero siempre con el aval y consentimiento de la comunidad''. Esta diversidad demuestra que existe la necesidad de que los investigadores tengan una relación cercana con las comunidades para entender los antecedentes y los objetivos de cada caso en particular.

Como se discutió anteriormente, la calidad del sistema final es una preocupación importante para muchos participantes. En la Figura \ref{fig:results}(f) podemos ver que publicar un sistema MT experimental también es controvertido. La posibilidad de utilizar un sistema experimental gusta a $54,8\%$ de nuestros participantes, que es ligeramente superior al número de participantes que están en contra ($45,5\%$). Algunas opiniones en contra están en línea con preocupaciones anteriores sobre traducciones incorrectas de contenido cultural: ``Algo que está desprovisto de estructura y objetividad cultural no es
no se puede poner a disposición del público'' y ``... se causaría daño a la lengua y sus representantes ya que los estudiantes aprenderían de forma incorrecta''. La mayoría de las personas con una opinión positiva están de acuerdo en que un sistema inicialmente pobre podría mejorarse con el tiempo: ``Si pudiera mejorarse y corregirse, sería excelente''.

\subsection{Discusión}

En la Sección \ref{sec:theory} examinamos el debate en curso sobre la ética en la documentación, la traducción, antes de presentar un estudio de entrevista en la Sección \ref{sec:study}. Ahora discutimos algunos de los problemas más importantes que hemos identificado en la última sección con más profundidad.

\paragraph{Necesidad de consultas con las comunidades} Experiencias anteriores \cite{bird-2020-decolonising,liu2022not} así como nuestro estudio resaltan la necesidad de consultar con las comunidades indígenas cuando se realizan investigaciones relacionadas con sus idiomas\footnote{Un ejemplo de un trabajo de campo comprometido con la comunidad es \newcite{czaykowska2009research}}. En algunos casos, el requisito mínimo expresado es informar a los ponentes sobre los nuevos avances tecnológicos. La retroalimentación y los controles de calidad también son cruciales para los sistemas de traducción automática e importantes para los miembros de las comunidades. Esta consulta debe incluir el diálogo intercultural ya que ha sido un instrumento central en la toma de decisiones de las comunidades indígenas \cite{beauclair2010eticas}. Recomendamos hacer esto integrando a los miembros de la comunidad en el ciclo y, por supuesto, dándoles el crédito que se merecen.

\paragraph{Sistemas legales versus visiones tradicionales de la propiedad del conocimiento comunal} Los sistemas legales y, con eso, las leyes de derechos de autor varían según el país. Sin embargo, las leyes a veces están en conflicto con los puntos de vista tradicionales de los pueblos indígenas \cite{dwyer2006ethics}. Por lo tanto, cuando se trabaja con comunidades indígenas, recomendamos discutir y acordar los derechos de propiedad con los anotadores o los participantes del estudio antes de comenzar el trabajo para encontrar un arreglo con el que todos estén contentos. También nos gustaría señalar que, según nuestro estudio de caso, la sensación general es que los datos y los resultados de la investigación deben ser accesibles para la comunidad que habla el idioma. Esto contradice la práctica de algunos esfuerzos de documentación que cierran los datos recopilados al público e incluso a los hablantes de la comunidad \cite{Heriberto2021Nuevas}. Algunos participantes en nuestro estudio incluso sugieren el uso de Creative Commons (CC)\footnote{\url{https://creativecommons.org/licenses/}} para los datos. Sin embargo, el uso de CC podría no ser la mejor opción de licencia, ya que no está diseñado específicamente para las necesidades de los indígenas.
Finalmente, siempre que los datos recopilados se utilicen para uso comercial, los acuerdos especiales que involucran aspectos financieros son cruciales.

\paragraph{Permisos} Algunas comunidades requieren que se obtenga un permiso de su entidad gobernante cuando alguien, que no es miembro, quiere estudiar su idioma. Esto puede ser difícil ya que a veces no existe una autoridad central. Averiguar de quién obtener el permiso puede ser un desafío en tales escenarios. Sin embargo, como vemos en este estudio, muchas comunidades no requieren este permiso. Un proyecto prometedor que pretende simplificar este tema son las etiquetas KP\footnote{
\url{https://localcontexts.org/labels/tradicional-conocimiento-etiquetas/}}. Es un conjunto de etiquetas que las comunidades pueden usar para expresar sus permisos y voluntad de cooperar con investigadores y proyectos externos.

\paragraph{Datos personales} De las respuestas abiertas, aprendemos además que, para muchos hablantes, usar su propio idioma en su entorno diario les ayuda a proteger su privacidad:
Sus conversaciones solo pueden ser entendidas por su familia o entorno cercano. Sin embargo, esta preocupación por el manejo de datos también es válida para otros lenguajes.

\paragraph{Preocupaciones sobre
Información Privada de la Comunidad} El punto anterior puede extenderse aún más a las asambleas y otras reuniones organizacionales, donde se utiliza la barrera del idioma para mantener en privado sus decisiones o estrategias. Esta es una preocupación que tienen las comunidades con MT y los posibles temas que podrían ser perjudiciales para ellos. Algunas comunidades también tienen preocupaciones generales acerca de compartir su idioma con personas ajenas a la comunidad (por ejemplo, la controversia del Diccionario Hopi \cite{hill2002publishing}). Para este caso, es importante no abordar este tema desde un punto de vista legal occidental e ir hacia las prácticas y normas tradicionales de gobierno interno y la consulta con las comunidades.

\paragraph{La religión y la Biblia} Con respecto a los dominios problemáticos para MT, varios participantes de la encuesta mencionaron la religión. Esto es bastante relevante para la comunidad de NLP, ya que el recurso más amplio disponible actualmente para los idiomas minoritarios es la Biblia. Como se vio en la Sección \ref{subsec:ethics_translation}, el uso colonial de la traducción de textos religiosos \cite{niranjana1990translation} es precisamente el origen de estos conflictos. Por lo tanto, recomendamos que los investigadores de NLP y MT usen la Biblia con cuidado, a través de un proceso de consulta, y consideren sus impactos. Sin embargo, sin una relación cercana con cada comunidad (por ejemplo, en un experimento masivo de MT multilingüe), la recomendación es evitar el uso de la Biblia.

\paragraph{Tecnología y soberanía de los datos} Tener tecnología para sus propios idiomas es bien visto por la mayoría de los participantes del estudio. Sin embargo, también encontramos un fuerte deseo de participar directamente en el desarrollo de sistemas MT. Esto requiere una mayor inclusión de investigadores indígenas en la PNL. Por lo tanto, la formación de investigadores e ingenieros indígenas es una tarea importante que recomendamos sea valorada más por las comunidades de NLP y MT. Somos conscientes de que las desigualdades existentes no se pueden eliminar de inmediato o de forma aislada, pero todos pueden brindar su apoyo.\footnote{La soberanía tecnológica es un tema central para la conferencia Natives in Tech en 2022: \url{https://nativesintech.org/conference /2022}} La creación de un proceso colaborativo es una propuesta que surge de las propias comunidades: ``Tecnología como Tequio; la creación tecnológica y la innovación como bien común'' \cite{aguilar2020}. Sin embargo, no es posible construir tecnologías contemporáneas de NLP sin datos. Y esto abre la discusión sobre la Soberanía de los Datos. En primer lugar, es importante mencionar que las comunidades tienen derecho a la autodeterminación, y esto incluye los datos que generan. Aplicar esta soberanía a los datos se refiere a tener control sobre los datos, el conocimiento\footnote{Ver \url{https://indigenousinnovate.org/downloads/indigenous-knowledges-and-data-governance-protocol_may-2021.pdf}} y cultural. expresiones que son creadas por estas comunidades. Como se discute en este documento, es importante llegar a acuerdos con las comunidades a través de consultas y colaboraciones directas. Esto incluye la concesión de licencias y la propiedad de los productos de datos finales.

\paragraph{Nuestros hallazgos y trabajos anteriores} Finalmente, queremos relacionar nuestros hallazgos con discusiones similares en trabajos anteriores. La mayoría de las inquietudes y sugerencias anteriores relacionadas con la inclusión y consulta de personas de las comunidades \cite{bird-2020-decolonising,liu2022not} están alineadas con los deseos y anhelos de los participantes en nuestro estudio. La inclusión de miembros de la comunidad como coautores \cite{liu2022not} no debe ser una mecánica artificial, sino más bien un proceso de inclusión amplio, que incluye la soberanía de los datos y la tecnología. Esto también está alineado con la construcción de la comunidad a la que apunta \citet{zhang2022can}. Además, debemos considerar que pueden existir temas problemáticos y no subestimar la importancia de las traducciones de alta calidad.

\subsection{Conclusión}
En este trabajo, que se centra en los desafíos éticos para la MT de las lenguas indígenas,
primero brindamos una descripción general de los enfoques éticos relevantes, los desafíos éticos para la traducción en general y los desafíos más específicos para la traducción automática. Luego, llevamos a cabo un estudio de caso, para el cual entrevistamos a activistas de lenguas indígenas, profesores de idiomas y líderes comunitarios de las Américas.
Nuestros hallazgos se alinearon con hallazgos previos con respecto a la necesidad de inclusión y consulta con las comunidades cuando se trabaja con datos lingüísticos. Además, nuestros participantes expresaron un interés sorprendentemente fuerte en tener sistemas de traducción automática para sus idiomas, pero también preocupaciones sobre el uso comercial, el mal uso cultural y religioso, los datos y la soberanía tecnológica. Terminamos con recomendaciones específicas para las comunidades de NLP y MT, pero aún más importante, un marco de discusión abierto para las comunidades indígenas.

\end{document}